\documentclass[10pt, twocolumn, letterpaper]{article}

\usepackage[pagenumbers]{cvpr}      

\usepackage{float}
\usepackage{bbding}
\usepackage{pifont}
\usepackage{graphicx}
\graphicspath{{./res/figs/}}
\usepackage{amsmath}
\usepackage{amssymb}
\usepackage{comment}
\usepackage{booktabs}
\usepackage{multirow}
\usepackage{makecell}
\usepackage{titletoc}
\usepackage{setspace}
\usepackage{appendix}
\usepackage[table]{xcolor}
\usepackage[hang,flushmargin]{footmisc}

\usepackage{color, colortbl}
\definecolor{Gray}{gray}{0.9}

\usepackage[pagebackref, breaklinks, colorlinks]{hyperref}

\usepackage[capitalize]{cleveref}
\crefname{section}{Sec.}{Secs.}
\Crefname{section}{Section}{Sections}
\Crefname{table}{Table}{Tables}
\crefname{table}{Tab.}{Tabs.}

\makeatletter
\newcommand\figcaption{\def\@captype{figure}\caption}
\newcommand\tabcaption{\def\@captype{table}\caption}
\makeatother

\begin{document}

\title{Learning Open-vocabulary Semantic Segmentation Models \\ From Natural Language Supervision}

\author{Jilan Xu$^{1,2}$ ~Junlin Hou$^{1}$ ~Yuejie Zhang$^{1}$ ~Rui Feng$^{1}$ ~Yi Wang$^{2}$ Yu Qiao$^{2}$ ~Weidi Xie$^{2,3}$
\\  [2pt]
$^1$School of Computer Science, Shanghai Key Lab of Intelligent Information Processing, \\ [2pt]
Shanghai Collaborative Innovation Center of Intelligent Visual Computing, Fudan University\\  [2pt]
$^2$Shanghai AI Laboratory
$^3$Shanghai Jiaotong University \\ [2pt]
\url{https://jazzcharles.github.io/OVSegmentor/}
}

\maketitle

\begin{abstract}
In this paper, we consider the problem of open-vocabulary semantic segmentation (OVS), which aims to segment objects of arbitrary classes instead of pre-defined, closed-set categories. The main contributions are as follows: 
\textbf{First}, we propose a transformer-based model for OVS, termed as OVSegmentor, which only exploits web-crawled image-text pairs for pre-training without using any mask annotations. OVSegmentor assembles the image pixels into a set of learnable group tokens via a slot-attention based binding module, and aligns the group tokens to the corresponding caption embedding. 
\textbf{Second}, we propose two proxy tasks for training, namely masked entity completion and cross-image mask consistency. The former aims to infer all masked entities in the caption given the group tokens, 
that enables the model to learn fine-grained alignment between visual groups and text entities. The latter enforces consistent mask predictions between images that contain shared entities, which encourages the model to learn visual invariance. 
\textbf{Third}, we construct CC4M dataset for pre-training by filtering CC12M with frequently appeared entities, which significantly improves training efficiency.
\textbf{Fourth}, we perform zero-shot transfer on three benchmark datasets, PASCAL VOC 2012, PASCAL Context, and COCO Object. Our model achieves superior segmentation results over the state-of-the-art method by using only 3\% data (4M vs 134M) for pre-training. Code and pre-trained models will be released for future research.
\end{abstract}


\section{Introduction}
\label{sec:intro}

Semantic segmentation considers the problem of assigning semantic labels to each pixel in the image. It plays central roles in a wide range of real-world scenarios, including autonomous driving, computer-aided diagnosis and satellite image analysis, to name a few. 
Generally speaking, two lines of research dominate semantic segmentation, one idea is to cluster the pixels into different groups and assign a semantic label to each group; the other idea treats segmentation as pixel-wise classification, casting each of the pixels into one category. Despite tremendous progress, the scalability of existing approaches that rely on supervised training has been fundamentally limited: 
(1) costly annotation procedure. Extensive manual pixel-wise annotations are required for training segmentation models; 
(2) closed-set segmentation. The model is restricted to segmenting objects from a closed-set of categories. Whenever a new dataset comes, the model requires fully-supervised re-training.



\begin{figure}[t]
\centerline{\includegraphics[width=\columnwidth]{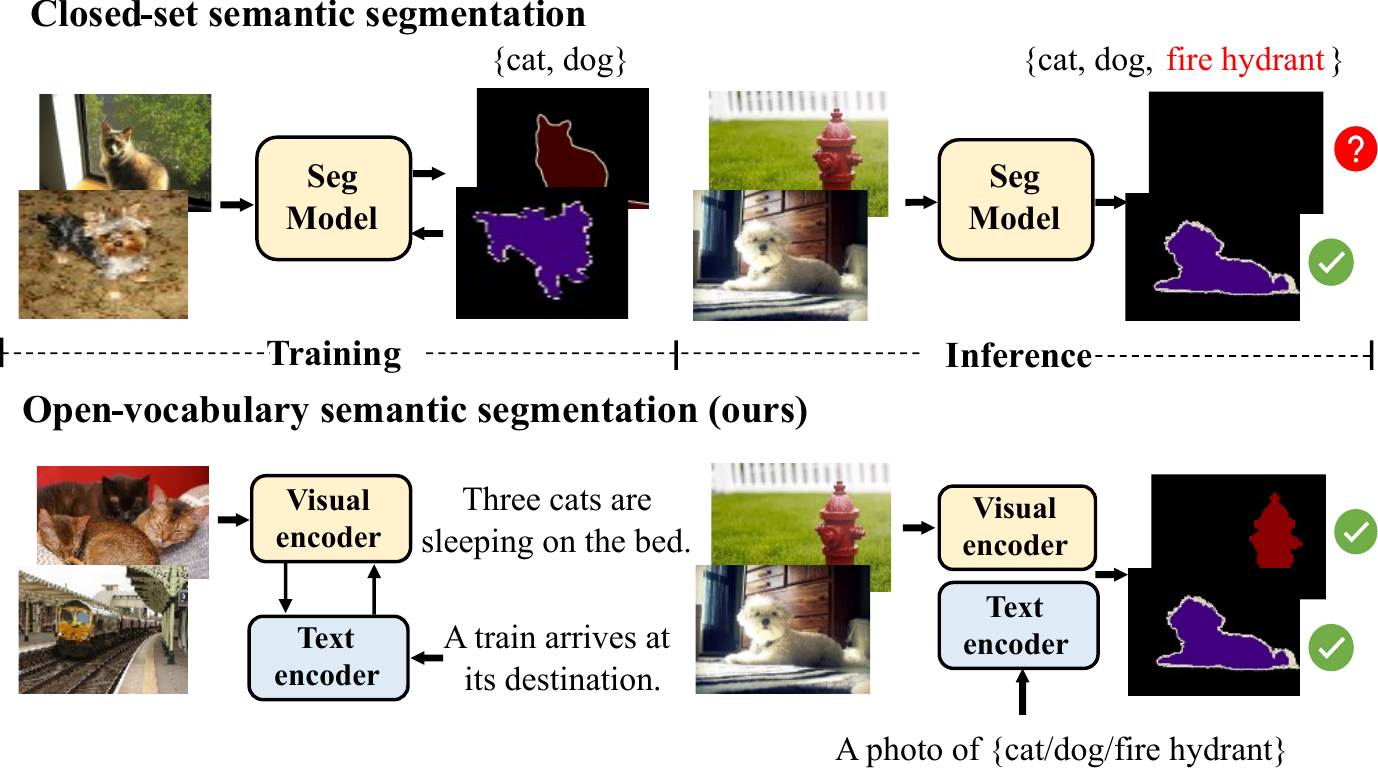}}   
\vspace{-5pt}
\caption{\textbf{An illustration of open-vocabulary semantic segmentation.} Models trained on closed-set classes (cat and dog) fail to segment novel class (fire hydrant). We train a visual encoder and a text encoder on web-collected image-text pairs without using any mask labels, and our model can segment arbitrary object classes.}
\vspace{-0.5cm}
\label{fig:intro}
\end{figure}

In this paper, our goal is to train an open-vocabulary semantic segmentation (OVS) model, by exploiting the freely available image-caption pairs on Internet, as illustrated in Fig.~\ref{fig:intro}. The recent CLIP~\cite{clip} and ALIGN~\cite{align} papers have demonstrated that a combination of large-scale image-caption pairs, with simple noise contrastive estimation can be used to learn powerful image-text embeddings from scratch, and show strong ``zero-shot” generalization abilities for open-vocabulary classification. Recent works, such as GroupViT~\cite{gvt}, initially extend the idea towards semantic segmentation by training a segmentation model with text supervision only. They perform hierarchical grouping of visual tokens, which are then aligned to the corresponding text embeddings via a contrastive loss. However, the following issues remain challenging and unsolved: \textit{First}, the captions only provide coarse, image-level descriptions, which are insufficient for training semantic segmentation models where fine-grained, pixel-wise supervision is usually needed. 
\textit{Second}, the diversity of web-collected data is large, that requires the model to learn visual invariance on objects of interest, with only weak supervision provided. For instance, the visual appearance of two images with similar captions can be drastically different.



To tackle the above challenges, 
(\romannumeral1) we propose a transformer-based model for open-vocabulary semantic segmentation, dubbed as OVSegmentor, 
that can segment objects of arbitrary categories via zero-shot transfer, with only image-caption pairs for pre-training.
Specifically, we introduce learnable group tokens to cluster image patches via a slot-attention~\cite{slot} based binding module, 
and align the group tokens with the corresponding caption embedding. Note that our model neither needs ground-truth masks for training nor requires additional re-training on target segmentation datasets, substantially alleviating the annotation efforts and boosting transfer efficiency; 
(\romannumeral2) As for training on the image-caption dataset, we propose two proxy tasks, namely masked entity completion and cross-image mask consistency, based on an observation that entities in the captions matter in matching specific visual objects. The former trains the model to infer \textit{all} the masked entities in the sentence given the group tokens, and the latter enforces consistent mask prediction for images with the common entity. Both tasks have shown to be beneficial in learning entity-specific, fine-grained and visually invariant group semantics;
(\romannumeral3) We construct an image-caption dataset, termed as CC4M, by designing an automatic approach to filter CC12M~\cite{cc12m} with frequently appeared informative entities, significantly improving the training efficiency.



We pre-train the proposed OVSegmentor on our filtered image-caption dataset~(CC4M), and manual segmentation masks are {\em not} used whatsoever. The model is evaluated on three segmentation benchmarks, PASCAL VOC 2012~\cite{voc}, PASCAL Context~\cite{context} and COCO~\cite{coco} in a zero-shot manner, 
{\em i.e.}, the model is directly evaluated on downstream benchmarks without any finetuning. 
Extensive experiments demonstrate that our model surpasses the model with supervised finetuning and outperforms state-of-the-art methods on PASCAL VOC by using only 3\% data (4M vs 134M) for pre-training, significantly improving the training efficiency.

\section{Related Work}
\vspace{2pt}
\noindent \textbf{Vision-Language Pre-training.} Vision-language pre-training (VLP) aims to learn joint visual-textual representations for a variety of multimodal downstream tasks. Existing works either learn unimodal encoders by distinguishing the positive pair(s) from the unpaired samples \cite{clip,frozen,loupe} or focus on one multimodal encoder for joint feature learning with masked image/language modeling and image-text matching losses \cite{uniter,alberf,alpro}. 
Additionally, some approaches seek fine-grained supervision for cross-modal interaction \cite{glip,glipv2,filip,xvlm,bridge}. For example, GLIP \cite{glip} proposed to align the bounding boxes with corresponding phrases in the text. However, they still rely on ground-truth grounding annotations. In contrast, our work explores fine-grained information with only weak supervision provided. 
Despite remarkable performance on multimodal downstream tasks, few of these vision-language models have been designed for fundamental vision tasks ({\em e.g.}, semantic segmentation).

\vspace{2pt}
\noindent \textbf{Zero-shot/Open-vocabulary Semantic Segmentation.}
The goal of zero-shot semantic segmentation is to segment objects of interest that are not seen in the training set. Prior works \cite{spnet,gagnet,strict} mainly transfer the knowledge from the training set (seen) to the testing set (unseen) via visual-semantic mapping. Inspired by the open-vocabulary nature of language, current approaches \cite{lseg,openseg,ma2022open,maskclip,zegformer,reco,namedmask} exploit vision-language models ({\em e.g.}, CLIP \cite{clip}) pre-trained on large-scale image-caption pairs. They need either finetuning \cite{lseg,openseg,zegformer} or self-training \cite{maskclip} on the target segmentation dataset, which is less efficient and flexible than zero-shot transfer. VLP for open-vocabulary segmentation combines the merits of both zero-shot transfer and open-vocabulary recognition. GroupViT \cite{gvt} designed a grouping vision transformer and learned the alignment between groups and text via the contrastive loss. ViL-Seg \cite{vilseg} combined image-text contrastive loss with online pixel clustering for segmentation. A concurrent work CLIPpy \cite{clippy} explored different aggregation operations for training spatial-aware vision-language models for segmentation. Beyond the global image-text matching, we further design masked entity completion and cross-image mask consistency to enrich the group semantics. 

\vspace{2pt}
\noindent \textbf{Fully-/Weakly-/Semi-Supervised Semantic Segmentation.} Fully-supervised semantic segmentation emerges from per-pixel classification \cite{fcn,deeplab} to mask classification \cite{maskformer,maxdeeplab}. To relieve the laborious annotations, extensive efforts have been made to address semantic segmentation with less supervision. 
In the family of weakly-supervised object localization \cite{bam,cream,kim2022bridging} and semantic segmentation \cite{ahn2018affinity,xie2022c2am,chen2022self}, only class labels are available for supervision. Generally, the class activation maps \cite{cam} derived from the classification network serve as the initial segmentation results. 
Another line of research focuses on semi-supervised semantic segmentation \cite{french2019semi,luo2020semi,ouali2020semi,chen2021semi,wang2022semi} where a few samples have dense per-pixel labels and the remaining samples are unlabeled. These works mainly perform supervised training on labeled samples with additional consistency regularizations posed on unlabeled samples. 
Despite promising results, these approaches are still limited to closed-set object categories. 


\begin{figure*}[t]
\centerline{\includegraphics[width=.98\textwidth]{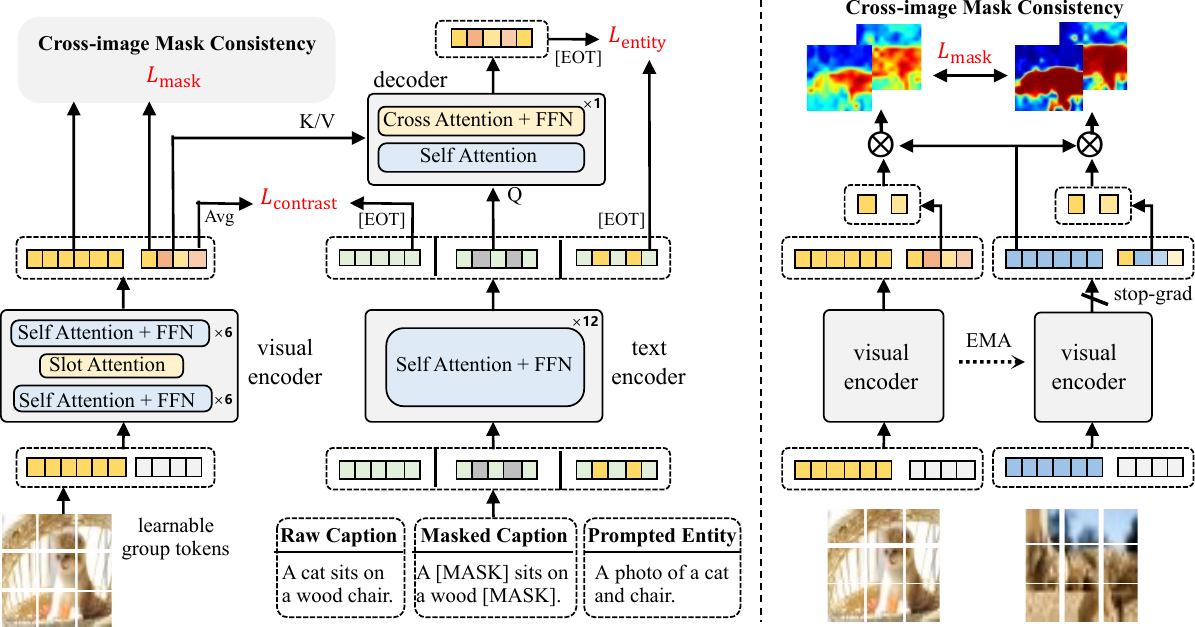}}
\vspace{-5pt}
\caption{\textbf{Overall framework.} The model takes an image-text pair as input, and outputs the image features and a set of learnable group tokens, which is aligned to the caption embedding via $\mathcal{L}_{\text{contrast}}$. The visual encoder consists of Transformer encoder layers, with a slot-attention based binding module in between to cluster the image tokens into groups. To enrich group semantics, (1) $\mathcal{L}_{\text{entity}}$ trains the model to infer \textit{all} masked entities in the caption given the group tokens. (2) $\mathcal{L}_{\text{mask}}$ enforces consistent mask predictions of images with the same entity (right figure). During inference, only visual and text encoders and learned group tokens are used to generate the segmentation mask.}
\vspace{-5pt}
\label{fig:model}
\end{figure*}

\section{Methodology}
Assuming we are given a dataset of image-caption pairs, 
{\em i.e.}, $\mathcal{D}_{\text{train}} =  \{(I_1, T_1), \dots, (I_N, T_N)\}$, where $I_i \in \mathbb{R}^{H\times W \times 3}$ denotes the image, and $T_i$ refers to its corresponding alt-text downloaded from the Internet, open-vocabulary semantic segmentation (OVS) is to train a semantic segmentation model only with these (very) weak labels, that can segment objects of arbitrary classes.

We present OVSegmentor, a vision-language pre-training framework for OVS. It assembles the image patches into groups and aligns the group to human-understandable categories, by only exploiting weak supervisions from web-collected image-caption pairs. Conceptually, the architecture consists of two stages, namely, a pixel-to-group binding that assigns all pixels with same semantics into one group, 
and group-to-category alignment that computes matching scores between each of these group tokens with semantic categories, the segmentation mask $m$ can be computed as:
\begin{align}
    m = \mathop{\arg\max}_C \mathbb{A} \cdot \mathcal{G} \cdot \mathcal{T}^\top ,
\label{eq:mask}
\end{align}
where $\mathbb{A}\in\mathbb{R}^{HW\times K}$ characterises a soft binding procedure, denoting the likelihood of each image pixels being assigned into $K$ learnable group embeddings~($\mathcal{G} \in \mathbb{R}^{K\times D}$); and $\mathcal{T} \in \mathbb{R}^{C\times D}$ denotes the embeddings with dimension $D$ for a total of $C$ categories, 
computed from a text encoder.
{\bf Note that}, at training time, manual segmentation masks are not present, and the intermediate binding and assignment can only be learnt from the image-caption pairs implicitly.

At inference stage, the pixel-group affinity $\mathbb{A}$ can be computed by the product between the image features $\mathcal{I}\in\mathbb{R}^{HW\times D}$ and the learned group tokens $\mathcal{G}$, and $\mathcal{T}$ can be derived by encoding the prompted class labels ({\em e.g.}, A photo of \{class\}).
Thus the mask prediction can be calculated via Eq.~\ref{eq:mask}. In the following sections, we detail the overall architecture and the training procedure for acquiring the components. The overall framework is illustrated in Fig.~\ref{fig:model}.



\subsection{Architecture}
\label{sec:architecture}
In this section, we will detail the proposed architecture at training stage, including a visual encoder for learning group tokens and the binding procedure, a text encoder that computes variants of caption embeddings. We describe the encoding procedure for single image and caption pair, the subscripts are thus ignored for simplicity.


\vspace{-6pt}
\subsubsection{Visual Encoder} 
\label{sec:visenc}
Given one training image-caption pair $(I, T)$,
we first split the image into $L=HW / P^2$ non-overlapping patches with patch size $P$. 
These patches are then transformed into a sequence of image tokens with MLPs, which we reuse
$I\in \mathbb{R}^{L\times D}$ to represent them for the ease of understanding.
Additionally, we introduce $K$ learnable tokens $G\in \mathbb{R}^{K\times D}$, 
that aims to group the image tokens into clusters. 

The visual encoder consists of two components, namely, Transformer encoders and binding modules. Specifically, the image tokens and learnable group tokens are concatenated, 
and iteratively processed by the Transformer encoders, and a binding module with slot-attention \cite{slot} being adopted for grouping. 
The visual encoder is defined as:
\begin{align}
    [\mathcal{G}; \mathcal{I}] &= \Phi_{\text{enc}}^2(\cdot)\circ \Phi_{\text{bind}}(\cdot) \circ \Phi_{\text{enc}}^1([G; I]),
\end{align}
where $\Phi_{\text{enc}}^1(\cdot)$ and $\Phi_{\text{enc}}^2(\cdot)$ refer to modules with Transformer encoder layers, and $\Phi_{\text{bind}}(\cdot)$ is the binding module. 
The output $\mathcal{G}\in \mathbb{R}^{K\times D}$ is the encoded group tokens,
and $\mathcal{I}\in \mathbb{R}^{L\times D}$ refers to the output image tokens.




\vspace{3pt}
\noindent\textbf{Transformer Encoder.} 
Both $\Phi_{\text{enc}}^1$ and $\Phi_{\text{enc}}^2$ consist of 6 Transformer encoder layers \cite{transformer}, where each layer is composed of a multi-head self-attention (MHSA) layer followed by layer normalisation (LN) and a feed-forward network (FFN). 
$\Phi_{\text{enc}}^1$ takes the concatenation of the image patches and the randomly initialised group tokens as input, and outputs intermediate encoded group and image tokens~($G'$ and $I'$); 
$\Phi_{\text{enc}}^2$ processes the output from the binding module.

\vspace{3pt}
\noindent\textbf{Binding Module.} 
The binding module uses slot-attention to cluster image tokens into groups in a data-dependent manner, {\em i.e.}, image patches with similar appearance and semantics are encouraged to be grouped together. Formally, the binding module accepts the output from the Transformer encoder $\Phi_{\text{enc}}^1$,
and transforms them into query, key and values with linear transformations:
\begin{align}
    \textbf{Q} = W_q^{\text{bind}}G', \quad \textbf{K}, \textbf{V} = W_k^{\text{bind}}I', \text{ } W_v^{\text{bind}}I'.
\end{align}
In contrast to the standard cross attention in Transformer Decoders~\cite{transformer}, 
slot-attention performs normalisation over queries, 
encouraging each image token to be claimed by one of the group tokens. 
The binding process can be defined as:
\begin{align}
    \mathbb{A}_{j,k} = \frac{\exp(\textbf{K}_j\cdot\textbf{Q}_k)}{\sum_{l}\exp(\textbf{K}_j\cdot\textbf{Q}_l)}, 
\label{eq:affinity}
\end{align}
where $\mathbb{A}_{j,k}$ refers to the probability of the $j$-th image token being assigned to the $k$-th group. Then, each updated group token $\tilde{G}'_k$ is computed by taking the weighted average of the image tokens that are assigned to it. The output of the binding module $G^{\text{bind}}$ is defined as:
\begin{align}
     \tilde{G}'_k &= \frac{\sum_{j=1}^{L}\mathbb{A}_{j,k}\textbf{V}_j}{\sum_{j=1}^{L} \mathbb{A}_{j,k}}, \\
     G^{\text{bind}} &= G' + W_o^{\text{bind}} \tilde{G}',
\end{align}
where $W_o^{\text{bind}}$ is a linear transformation.
Now we have obtained the correspondence between each pixel and group tokens, 
next we describe the procedure for encoding captions.

\vspace{-6pt}
\subsubsection{Text Encoder}
\label{sec:textenc}
Till this end, we start by filtering all the captions and only keep the ones with informative entities, followed by exploiting three variants of the caption encoding,
namely, the entire caption embedding, the masked caption embedding and the prompted entity embedding. In all cases we use a pre-trained BERT~\cite{bert} as the text encoder $\Phi_{\text{text}}$. 

\vspace{3pt}
\noindent \textbf{Constructing Entity Set.}
We adopt the nltk toolkit~\cite{nltk}
to extract entities from all the captions, and construct an entity set $\Omega = \Phi_{\text{entity}}(\{T_1, \dots, T_N\}) $ that only maintains the frequently appeared entities ({\em e.g.}, people, cat, shirt, etc.), and exclude the abstract nouns ({\em e.g.}, art, view, etc) as they usually do not correspond to any specific region in the image.
For each image-caption pair, 
we can thus obtain an image-caption-entity triplet $(I, T, E)$, 
where $E = \{ e | e \in T\enspace\cap\enspace e\in\Omega \}$ includes \textit{all} frequent entities in the caption. 


\vspace{3pt}
\noindent\textbf{Caption Embedding.}
Here, for each caption $T$, it is tokenised using the BERT tokeniser \cite{bert}, 
with the start-of-token [SOT] and end-of-token [EOT] added to the begining/ending. 
The caption embedding is computed as $\mathcal{T}^{\text{cap}} = \Phi_{\text{text}}(T) \in \mathbb{R}^{M \times D}$,
where $M$ is length after tokenisation. 

\vspace{3pt}
\noindent\textbf{Masked Caption Embedding.} 
In the second variant, we mask \textit{all} the frequent entities in the caption,
and compute the masked caption embedding as $\mathcal{T}^{\text{m-cap}} = \Phi_{\text{text}}(g(T)) \in \mathbb{R}^{M \times D}$, where
$g(\cdot)$ denotes an masking operation, that replaces the entity with a special [MASK] token. 

\vspace{3pt}
\noindent\textbf{Prompted Entity Embedding.} 
We construct manual prompts for each of the entities in the triplets, 
and compute its textual embedding as $\mathcal{T}^{\text{entity}} = \Phi_{\text{text}}(h(E)) \in \mathbb{R}^{M \times D}$,
where $h(\cdot)$ refers to the procedure for constructing manual prompts,
for example, we randomly sample a prompt template provided in \cite{clip}, 
and fill in the entities: \textit{A painting of a $\{\text{entity}_{1}\}$ and $\{\text{entity}_{2}\}$ and $\{\text{entity}_{3}\}$}. This sentence is padded to the same length as the caption after tokenisation. 


\subsection{Training}
As for training, we aim to learn the alignment between group tokens and caption embeddings via three proxy tasks, 
namely, image-caption alignment, masked entity completion, and cross-image mask consistency.

\vspace{3pt}
\noindent \textbf{Image-caption Alignment.} For each image-text pair, the objective is to align their visual and textual embeddings. The visual embedding $z^{\text{I}}$ is the average of group tokens, and the textual embedding $z^{\text{T}}$ is obtained by taking the [EOT] token feature of the caption embedding $\mathcal{T}^{\text{cap}}$, both projected to a 256-d joint feature space followed by normalisation. The image-caption contrastive loss $\mathcal{L}_{\text{contrast}}$ is formulated as:
\begin{align}
    \mathcal{L}_{\text{contrast}} = \frac{1}{2}\left(
    \frac{\exp({z^{\text{I}}_i}\cdot z^{\text{T}}_i)}{\sum_{l} \exp({z^{\text{I}}_i}\cdot  z^{\text{T}}_l)} + \frac{\exp({z^{\text{I}}_i}\cdot  z^{\text{T}}_i)}{\sum_{l} \exp({z^{\text{I}}_l}\cdot z^{\text{T}}_i)}\right).
\end{align}
Here, we omit the temperature parameter for simplicity.


\vspace{3pt}
\noindent \textbf{Masked Entity Completion.}
\label{sec:msc}
The goal of masked entity completion is to infer all the masked entities in the sentence given the group tokens. In specific, we adopt a Transformer decoder layer,
where a projection of the masked caption embedding is treated as query, 
and two linear transformations of group tokens are treated as key and values, respectively. 
\begin{align}
    \tilde{\mathcal{T}}^{\text{m-cap}} = \Phi_{\text{dec}}(W_q^{\text{dec}} \mathcal{T}^{\text{m-cap}},
      W_k^{\text{dec}} \mathcal{G}, 
      W_v^{\text{dec}} \mathcal{G}) \in \mathbb{R}^{M\times D},
\label{eq:dec}
\end{align}
where $W_q^{\text{dec}}$, $W_k^{\text{dec}}$ and $W_v^{\text{dec}}$ are linear transformations,
and $\tilde{\mathcal{T}}^{\text{m-cap}}$ denotes the updated vector sequence, with the masked entities being completed by querying group tokens.
For training, we extract the [EOT] token features $z^{\text{M}}$ and $z^{\text{E}}$ from $\tilde{\mathcal{T}}^{\text{m-cap}}$ and $\mathcal{T}^{\text{entity}}$, and construct a contrastive loss:
\begin{align}
    \mathcal{L}_{\text{entity}} = \frac{1}{2}\left(
    \frac{\exp({z^{\text{M}}_i}\cdot z^{\text{E}}_i)}{\sum_{l} \exp({z^{\text{M}}_i}\cdot  z^{\text{E}}_l)} + \frac{\exp({z^{\text{M}}_i}\cdot z^{\text{E}}_i)}{\sum_{l} \exp({z^{\text{M}}_l}\cdot z^{\text{E}}_i)}\right).
\label{eq:entity}
\end{align}
Intuitively, the entity completion task enables better alignment between the groups and entities. 


\vspace{3pt}
\noindent \textbf{Cross-image Mask Consistency.}
\label{sec:mcr}
To encourage visual invariance, we enforce consistent mask predictions between images that contain shared entities. Specifically, for each entity of interest, we can easily source multiple image-caption pairs from the dataset by text searching. Given one sampled image-caption-entity triplet, $(I_1,T_1, E_1)$, we can easily search for another sample $(I_2, T_2, E_2)$, with both triplet sharing one entity, {\em i.e.}, $e\in E_1 \cap E_2$. 
Given encoded groups $\mathcal{G}_1$, $\mathcal{G}_2$, two sets of subgroups $\mathcal{G}_1^{\text{sub}},\mathcal{G}_2^{\text{sub}}\in \mathbb{R}^{K'\times D}$ that represent the common entity $e$ ({\em e.g.}, cat in Fig.~\ref{fig:model}) are obtained by choosing groups with higher similarity to the entity embedding, where $K'=rK$ and $r$ is the selection ratio. 
The masks of co-attentive entity in image $I_1$ can be grounded by both entity-specific subgroups $\mathcal{G}_1^{\text{sub}},\mathcal{G}_2^{\text{sub}}$ as:
\begin{equation}
\mathcal{M}_{1}=\sigma(\mathcal{I}_1^\top\mathcal{G}_1^{\text{sub}}), ~
\hat{\mathcal{M}}_{1}=\sigma(\mathcal{I}_{1}^\top \mathcal{G}_{2}^{\text{sub}}) \in [0,1]^{L\times K'},
\end{equation}
where $\sigma$ is the sigmoid activation; both $\mathcal{M}_{1}=\{m_k^1\}_{k=1}^{K'}$ and $\hat{\mathcal{M}}_{1}=\{\hat{m}_k^1\}_{k=1}^{K'}$ consist of $K'$ unordered masks.

To align $\hat{\mathcal{M}}_{1}$ with $\mathcal{M}_{1}$, we first adopt the bipartite matching to find the optimal permutation $p^*_1$ over $K'$ subgroups with the lowest matching cost as:
\begin{equation}
    p^*_1 = \mathop{\arg\min}\limits_{p\in \mathcal{P}} \sum_{k} -\cos\left(m_k^1, \hat{m}_{p(k)}^1\right),
\label{eq:matching}
\end{equation}
where $\mathcal{P}$ is the full permutation and $\cos(\cdot)$ denotes the cosine similarity. Eq. \ref{eq:matching} is solved via the efficient Hungarian algorithm \cite{hungarian}. In this way, the symmetric cross-image mask consistency loss $\mathcal{L}_{mask}$ is defined as:
\begin{equation}
\small
    \mathcal{L}_{mask} = \frac{1}{2}
    \left(\sum_k \text{D}\left(\text{sg}(\textbf{m}^1_k),\hat{m}_{p^*_1(k)}^1\right) + \text{D}\left(\text{sg}(\textbf{m}^2_k),\hat{m}_{p^*_2(k)}^2\right)
    \right),
\end{equation}
where \text{sg}($\cdot$) denotes the stop gradient operation; the target mask $\textbf{m}_k$ is achieved by binarizing $m_k$ with a threshold $\delta$; 
$\text{D}(\textbf{m},\hat{m}) = 1 - 2|\textbf{m}\bigcap\hat{m}| / (|\textbf{m}|+|\hat{m}|)$ stands for the standard Dice loss. To guarantee the quality of pseudo mask targets, the masks are generated by an extra momentum model, which is updated by the exponential-moving-average (EMA) of the online model.

\vspace{3pt}
\noindent \textbf{Training Objective.}
We adopt a combination of three different loss functions:
\begin{align}
    \mathcal{L}_{\text{total}} = \mathcal{L}_{\text{contrast}} + \mathcal{L}_{\text{entity}}  + \lambda\mathcal{L}_{\text{mask}},
\end{align}
where $\lambda$ is the weight for balancing the mask consistency.




\subsection{Discussion}
\label{sec:disucssion}
One work that is closely related to ours is GroupViT \cite{gvt}, 
that improved the image-text alignment by exploiting nouns in the caption. In specific, they extracted multiple nouns and prompted each noun to a sentence to serve as extra matched captions for the image. The model is thus supervised by a multi-label contrastive loss. 
In contrast, 
our paper differs from GroupViT from three critical aspects: 
(1) Entities vs nouns. Rather than using all nouns, we leverage the entities that match to visual objects, enabling high-quality image-caption correspondence. 
(2) Network architecture. Beyond separate visual and text encoders in GroupViT, we further devise a (very) lightweight decoder to model the fine-grained, token-wise group-word correlation. 
(3) Proxy tasks for training. We propose two different proxy tasks, {\em i.e.}, masked entity completion and cross-image mask consistency to improve the entity-specific group semantics and further encourage visual invariance. The superiority of masked entity completion over multi-label contrastive loss is verified in Sec.~\ref{sec:ablation}.

\begin{table*}[t]
\centering
\resizebox{0.95\textwidth}{!}{
  \begin{tabular}{lccccccc}
    \toprule
    \multirow{2}{*}{Method}& 
    \multirow{2}{*}{Backbone} &
    \multirow{2}{*}{Pretrain dataset} & 
    \multirow{2}{*}{Supervision} & 
    \multirow{2}{*}{Zero-shot transfer} & 
    \multicolumn{3}{c}{Downstream datasets} \\
    \cmidrule(r){6-8}
    & & & & & PASCAL VOC & PASCAL Context & COCO Object \\
    \midrule
    DeiT \cite{deit} & ViT-S & IN-1K & class label & \XSolidBrush &  53.0 & 35.9 & - \\
    \midrule
    MoCo \cite{mocov3} & ViT-S & IN-1K & self & \XSolidBrush & 34.3 & 21.3 & - \\
    DINO \cite{dino} & ViT-S & IN-1K & self & \XSolidBrush & 39.1 & 20.4 & - \\
    MoCo \cite{mocov3} & ViT-S & CC12M+YFCC15M & self & \XSolidBrush & 36.1 & 23.0 & - \\
    DINO \cite{dino} & ViT-S & CC12M+YFCC15M & self & \XSolidBrush & 37.6 & 22.8 & - \\
    \midrule
    ViL-Seg \cite{vilseg} & ViT-B & CC12M & self+text  & \CheckmarkBold & 33.6 & 15.9 & - \\
    GroupViT \cite{gvt} & ViT-S & CC12M & text & \CheckmarkBold & 41.1 & - & - \\
    CLIPpy \cite{clippy} & ViT-B & CC12M &text & \CheckmarkBold & 50.8 & - & 23.8$^\dag$ \\
    GroupViT \cite{gvt} & ViT-S & CC12M+YFCC15M & text & \CheckmarkBold & 51.2 & \textbf{22.3} & 20.9 \\  \rowcolor{Gray}
    CLIPpy \cite{clippy} & ViT-B & HQITP-134M & text & \CheckmarkBold & 52.2 & - & 25.5$^\dag$ \\
    \midrule
    GroupViT* \cite{gvt} & ViT-S & CC4M & text & \CheckmarkBold & 19.8 & 8.8 & 9.1 \\
    GroupViT* \cite{gvt} & ViT-B & CC4M & text & \CheckmarkBold & 25.8 & 11.3 & 10.7 \\
    GroupViT* \cite{gvt} & ViT-S & CC12M & text & \CheckmarkBold & 40.2 & 18.7 & 17.7 \\
    \midrule 
    OVSegmentor (ours) & ViT-S & CC4M & self+text & \CheckmarkBold & 44.5 & 18.3 & 19.0 \\
    OVSegmentor (ours) & ViT-B & CC4M & self+text & \CheckmarkBold & \textbf{53.8} & 20.4 & \textbf{25.1} \\
    \bottomrule
  \end{tabular}}
\vspace{-5pt}
\caption{\textbf{Comparison with existing methods.} 
Models in the first five rows are finetuned on target datasets while the rest perform zero-shot transfer. GroupViT \cite{gvt} (with *) refers to our re-implementation.
Results of DeiT, MoCo, and DINO are copied from \cite{gvt}. Bold fonts refer to the best results among the models that enable zero-shot transfer. CLIPpy~\cite{clippy} uses 134M in-house data while our model uses an order of magnitude less data. $^\dag$ indicates evaluation on 133 COCO classes claimed by CLIPpy, which differs from ours (80 classes). }
\vspace{-0.4cm}
\label{tab:sota}
\end{table*}

\section{Experiments}
\subsection{Experimental Setups}
\label{sec: exps}
\noindent\textbf{Pre-training Dataset.} 
Following~\cite{gvt,vilseg}, we use Conceptual Captions 12M~\cite{cc12m} for training, which is originally constructed with over 12M image-text pairs collected from the Internet. However, due to some links have been expired, we have downloaded about 10M image-text pairs. 
The constructed entity set in Sec.~\ref{sec:textenc} includes a total number of 100 frequently appeared entities while abstract nouns ({\em e.g.}, art, view) are discarded. 
After filtering CC12M, we obtain 4.3 million image-text pairs for pre-training, which is termed as CC4M. 
Examples of entities include \textit{people}, \textit{car}, \textit{cup}, \textit{chair}, \textit{T-shirt}, \textit{house}, \textit{bed}, \textit{cat}, \textit{ball}, \textit{pizza}, {\em etc.} Please refer to the supplementary material for the full entity set. 


\vspace{2pt}
\noindent\textbf{Downstream Evaluation Datasets.}
We evaluate our model on three benchmarks, namely, 
PASCAL VOC 2012~\cite{voc}, PASCAL Context~\cite{context} 
and COCO Object~\cite{coco} with 20, 59 and 80 foreground classes, respectively. An extra background class is considered in all three datasets. We ignore their training sets and directly evaluate our method on the validation sets without any finetuning, 
including 1449, 5105, and 5000 images, respectively. 
In general, we report the mean Intersection-over-Union (mIoU) on all the classes.

\vspace{2pt}
\noindent\textbf{Implementation Details.} 
In our model, the self-attention layers in the visual encoder are initialised with DINO~\cite{dino} pre-trained on ImageNet. The text encoder is initialised with BERT~\cite{bert} model pre-trained on BookCorpus and English Wikipedia. Our decoder with one randomly initialised Transformer Decoder layer performs reasonably well. 
The input image is randomly cropped to 224$\times$224 at training time, and the batch size is set to 2048 with an initial learning rate $3.2\times10^{-4}$. We train our model for 40 epochs using the Adam optimizer with weight decay set to 0.5. The coefficient for updating the momentum model is 0.99.
As the generated masks are unreliable in early epochs, we set the mask consistency coefficient $\lambda$=0 for the first 30 epochs and $\lambda$=0.1 for the remaining epochs. The group selection ratio $r$ is 0.5.  
As for the threshold in mask consistency loss, we use $\delta$=0.65.
At inference time, the image is resized with a shorter length of 448. We follow~\cite{gvt} to set a threshold for the background class, which is $0.9$, $0.5$ and $0.9$ on PASCAL VOC, PASCAL Context and COCO Object, respectively.

\begin{figure*}[t]
\centerline{\includegraphics[width=\textwidth]{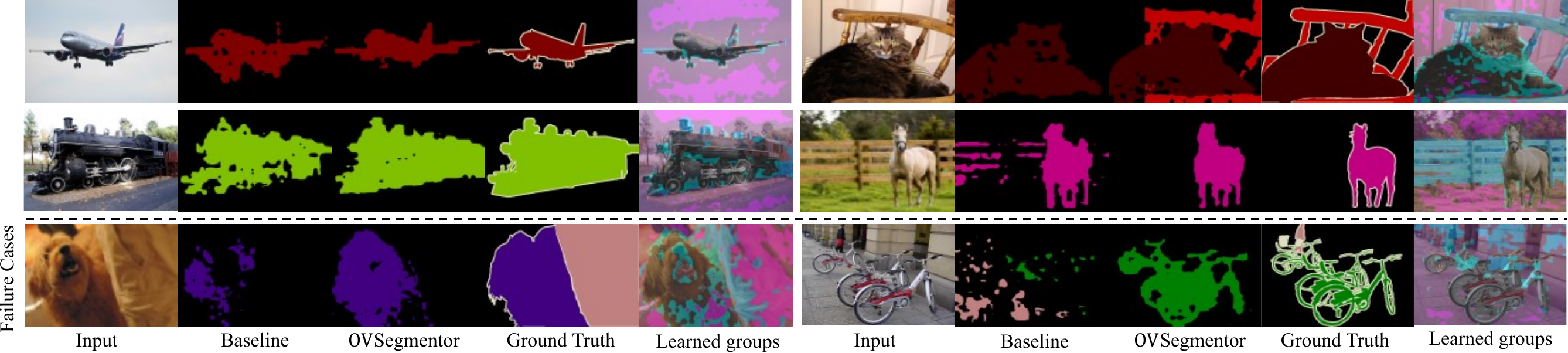}}  
\vspace{-8pt}
\caption{\textbf{Qualitative results on PASCAL VOC.} We show the learned groups in the 5\textsuperscript {th} and 10\textsuperscript{th} columns with each color representing a group. The last row shows two failure cases where the model fails to segment the person's leg (left) and the man at distance (right).}
\label{fig:visual}
\vspace{-8pt}
\end{figure*}

\subsection{Comparison with Existing Methods}
In Table~\ref{tab:sota}, we compare our model with the existing models that have been trained with (1) fully-supervised finetuning transfer and (2) zero-shot transfer.
In Table~\ref{tab:zero}, zero-shot segmentation (ZSS) approaches are listed for comparison.

\vspace{2pt}
\noindent\textbf{Comparison with Finetuning Transfer.} 
We compare our method with DeiT \cite{deit}, MoCo \cite{mocov3} and DINO \cite{dino},
which are pre-trained on ImageNet \cite{deng2009imagenet} or CC12M+YFCC15M datasets with class labels \cite{deit} or self-supervision \cite{mocov3,dino}, and finetuned on the training set from downstream benchmarks, with a randomly initialised convolution head appended on the backbone network. 
As shown in Table~\ref{tab:sota}, our model achieves competitive performance on PASCAL Context, and outperforms the self-supervised methods by over 10\% on PASCAL VOC, with zero-shot transfer.

\vspace{3pt}
\noindent\textbf{Comparison with Zero-shot Transfer.}
Here, we compare with existing works under the zero-shot transfer scenario, including GroupViT~\cite{gvt}, ViL-Seg~\cite{vilseg} and CLIPpy~\cite{clippy}, with the pre-training data ranging from CC12M~\cite{cc12m} to 134M in-house dataset HQITP-134M~\cite{clippy}. 
For fair comparison, we re-train GroupViT~\cite{gvt} with their official codebase on our CC4M and CC12M datasets, with the same pre-trained weights as ours being adopted on CC4M. However, we observe no further performance gain of either applying pre-trained weights to GroupViT or using ViT-B as the backbone on CC12M. The results on CC12M match the reported ones (40.2 vs 41.1). 
Under the same pre-trained ViT-B backbone and CC4M dataset, 
our method surpasses the original GroupViT by 35.7\%. 
Additionally, by only using 4M pre-training data, 
our model yields the best segmentation performance on PASCAL VOC,
even outperforming CLIPpy, which is a concurrent work to ours, 
and pre-trained on 134M data, indicating the effectiveness and training efficiency of our proposed model.


\begin{table}[t]
\centering
\resizebox{\columnwidth}{!}{
  \begin{tabular}{lccccc}
    \toprule
    Method& 
    Pretrain dataset& 
    Seen &
    Unseen & 
    VOC &
    Context \\
    \midrule
    SPNet \cite{spnet} & - & \CheckmarkBold &  \XSolidBrush  & 15.6 & 4.0  \\
    ZS3 \cite{z3net} & - & \CheckmarkBold &  \CheckmarkBold & 17.7 & 7.7  \\
    GaGNet \cite{gagnet} & -  & \CheckmarkBold &  \CheckmarkBold & 29.9 & 15.0 \\
    SIGN \cite{sign} & - & \CheckmarkBold &  \CheckmarkBold  & 28.9 & 14.9  \\
    Joint \cite{joint} & -  & \CheckmarkBold &  \XSolidBrush  & 32.5  & -  \\ \rowcolor{Gray}
    ZegFormer \cite{zegformer} & CLIP400M & \CheckmarkBold &  \XSolidBrush  & 63.6 &   \\ \rowcolor{Gray}
    MaskCLIP+ \cite{maskclip} & CLIP400M & \CheckmarkBold &  \XSolidBrush & 86.1 & 66.7$^\dag$ \\
    \midrule 
    ViL-Seg \cite{vilseg} & CC12M & \XSolidBrush & \XSolidBrush & 37.3 & 18.9 \\
    GroupViT \cite{gvt} & CC12M+YFCC15M & \XSolidBrush & \XSolidBrush  & 43.7  & 51.3 \\
    \midrule
    GroupViT* \cite{gvt} & CC4M &  \XSolidBrush & \XSolidBrush & 22.4 & 24.5  \\
    GroupViT* \cite{gvt} & CC12M &  \XSolidBrush & \XSolidBrush & 33.1 & 45.3 \\
    \midrule
    OVSegmentor & CC4M & \XSolidBrush & \XSolidBrush  & \textbf{46.6} & \textbf{54.5} \\
    \bottomrule
  \end{tabular}
}
\vspace{-5pt}
\caption{\textbf{Comparison with zero-shot segmentation methods on unseen classes.} Seen/Unseen denotes whether the model is trained on seen/unseen classes. Our method outperforms most zero-shot segmentation models even without training on seen classes. $\dag$ indicates a different set of unseen classes.}
\vspace{-0.4cm}
\label{tab:zero}
\end{table}

\vspace{2pt}
\noindent\textbf{Comparison with Zero-shot Segmentation Methods.} 
In this line of research~\cite{spnet,gagnet,zegformer}, 
the idea is to train the model with full mask labels~(obtained either from manual groundtruth or pseudo-labelling~\cite{maskclip}) on the seen classes and transfer the model to unseen classes, and the task is thus dubbed as zero-shot semantic segmentation (ZSS). 
To be specific, 5 classes (potted plant, sheep, sofa, train and tv-monitor) in PASCAL VOC and 4 classes (cow, motorbike, sofa and cat) in PASCAL Context are considered unseen while the remaining classes belong to seen. 
For comparison, we also report the zero-shot transfer performance of our model on these unseen classes,
however, {\bf note that}, we do not use any manual mask annotations for training. As observed in Table~\ref{tab:zero}, our model surpasses majority of the models trained under the ZSS scenario, except for~\cite{zegformer,maskclip} that adopted CLIP model pre-trained on 400M data. 
In terms of transfer efficiency, our model excels at zero-shot transfer ability without the need of training on seen classes.

\begin{table}
\centering
\resizebox{\columnwidth}{!}{
  \begin{tabular}{ccccc}
     \toprule
     Baseline & $\mathcal{L}_{\text{entity}}$ & $\mathcal{L}_{\text{mask}}$ & PASCAL VOC & PASCAL Context \\
     \midrule
      \CheckmarkBold &  &  & 40.5 & 15.1 \\
      \CheckmarkBold & \CheckmarkBold &  & 48.9 & 19.9 \\
      \CheckmarkBold &  & \CheckmarkBold & 44.8 & 17.0 \\
      \CheckmarkBold & \CheckmarkBold & \CheckmarkBold & 53.8 & 20.4 \\
     \bottomrule
  \end{tabular}
}
\vspace{-5pt}
\caption{\textbf{Ablation study} on the masked entity completion loss ($\mathcal{L}_{\text{entity}})$ and cross-image mask consistency loss ($\mathcal{L}_{\text{mask}}$). The baseline only uses the image-caption contrastive loss ($\mathcal{L}_{\text{contrast}}$).}
\vspace{-5pt}
\label{tab:ablations}
\end{table}


\subsection{Ablation Study}
\label{sec:ablation}
In this section, we conduct thorough ablation studies to validate the necessity of each proposed component.

\vspace{2pt}
\noindent\textbf{Ablation Study on Proxy Tasks.}
Here, we aim to understand the effects of our proposed proxy tasks, 
{\em i.e.}, masked entity completion and cross-image mask consistency. 
As shown in Table~\ref{tab:ablations}, the baseline model uses the image-text contrastive loss $\mathcal{L}_{\text{contrast}}$ only, while adding the entity completion task, $\mathcal{L}_{\text{entity}}$, we observe a significant improvement by 8.4\% and 4.8\% on PASCAL VOC and PASCAL Context, respectively. The performance gain is due to the ability of better aligning the pixel groups with the visual entities.
Additionally, the mask consistency also brings improvements, 
and combing both leads to the best performance. 
Qualitative results can be seen in  Fig.~\ref{fig:visual}. We refer the readers for more visualisations in the supplementary material.


\begin{figure*}[t]
\centerline{\includegraphics[width=\textwidth]{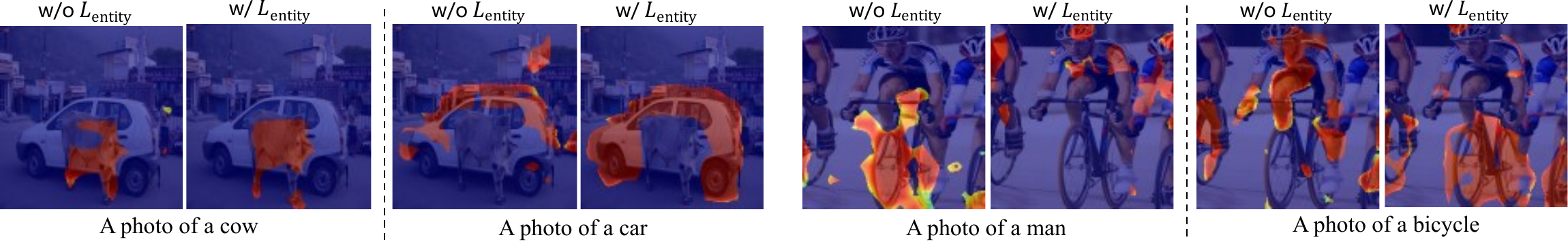}}   
\vspace{-8pt}
\caption{\textbf{Qualitative results of the masked entity completion loss.} We show different groups that match the given class. With the boost of group semantics, our model improves visual grouping (left) and solves false matching between group and text (right) of the baseline.}
\label{fig:mee}
\vspace{-8pt}
\end{figure*}

\begin{table}
\centering
\resizebox{0.95\columnwidth}{!}{
  \begin{tabular}{ccc}
     \toprule
     Masking objective & PASCAL VOC & PASCAL Context \\
     \midrule
     All entities (ours) & \textbf{48.9} & \textbf{19.9} \\
     Single entity & 47.0 & 18.5 \\
     All nouns & 45.4 & 17.4 \\
     Multi-label contrastive & 44.5 & 17.0 \\
     MLM (w/ groups) & 42.8 & 16.3 \\
     MLM (w/o groups) & 36.3 & 14.6 \\
     \bottomrule
  \end{tabular}
}
\vspace{-5pt}
\caption{\textbf{Comparison on masking objectives.} MLM (w/ groups) refers to masked language modeling \cite{bert} given the group tokens as keys and values in the decoder while MLM (w/o groups) directly predicts masked words from the output of text encoder.}
\vspace{-0.4cm}
\label{tab:maskobj}
\end{table}

\vspace{3pt}
\noindent\textbf{On the Choice of Masking Objectives.}
We compare our masked entity completion with a series of variants as shown in Table~\ref{tab:maskobj}. Our proposed objective of masking \textit{all entities} is listed in the first row. 
\textbf{(1) All entities vs one entity}: the masked language modeling (MLM) in prior works~\cite{bert,vilbert} normally choose 15\% of the token positions in the sentence for prediction, which results in one entity in most of our cases. 
We observe that masking \textit{all} entities is 2.9\% mIoU better than single entity masking, as it forces the network to infer all possible object categories in the image, that is potentially beneficial for the group to category alignment.
\textbf{(2) Entities vs nouns}: 
masking and predicting noun phrases in the sentence \cite{bridge} is one feasible option to learn fine-grained vision-text matching. However, noun masking leads to 3.5\% lower mIoU than our entity masking strategy. 
This is because not all of the nouns in the sentence are visually corresponding to the objects in the image ({\em e.g.}, illustration, night, etc), thus the group tokens are not expected to align with these nouns. Our method avoids this issue by only masking the visual entities. 
\textbf{(3) Masked entity completion vs multi-label contrastive loss:} comparing with the multi-label contrastive loss used in GroupViT~\cite{gvt}, our proposed strategy shows superior performance.
\textbf{(4) Masked entity completion vs masked language modeling}:
MLM \cite{bert} originally predicts the masked token over the entire vocabulary via a cross-entropy loss. 
Here, we restrict the vocabulary to our constructed entity set for fair comparison. 
Our masking strategy surpasses both MLM variants by a clear margin, which we conjecture is because: (1) MLM classifies each masked token \textit{individually}, and the model can easily refer to the context words without relating to groups. (2) MLM focuses on word-level representations, which is not in accordance with the sentence-level representation of the class embeddings we used during inference. 
Fig.~\ref{fig:mee} shows the effect of the masked entity completion in improving visual grouping (left) and group-text alignment (right).

\begin{table}[t]
\begin{minipage}{0.48\columnwidth}
\centering
\makeatletter\def\@captype{table}\makeatother
  \resizebox{1.0\columnwidth}{!}{
  \begin{tabular}{ccccc}
    \toprule
    $K$/$\tau$ & 0.25 & 0.5 & 0.75 & 1.0 \\
    \midrule
     8 & 49.0 & 53.6 & 51.8 & 50.4 \\
     16 & 42.3 & 43.2 & 43.1 & 42.9 \\
    \bottomrule
  \end{tabular}}
\vspace{-5pt}
\caption{Comparison on group numbers $K$ and selection ratios $\tau$ on VOC.}
\vspace{-5pt}
\label{tab:groupnumber}
\end{minipage}
\hfill
\begin{minipage}{0.48\columnwidth}
\centering
\makeatletter\def\@captype{table}\makeatother
\resizebox{1.0\columnwidth}{!}{
  \begin{tabular}{ccc}
    \toprule
    Objective & Loss & mIoU \\
    \midrule
    Mask Consistency & Dice & 53.8 \\
    Group Consistency & NCE & 47.9 \\
    \bottomrule
  \end{tabular}}
  \vspace{-5pt}
  \caption{Comparison on different objectives for cross-image consistency on VOC.}
  \vspace{-5pt}
  \label{tab:objective}
  \end{minipage}
\end{table}

\vspace{3pt}
\noindent\textbf{On Cross-image Mask Consistency.} 
Here, we analyze two key factors in cross-image consistency: 
(1) group numbers and selection ratios. 
As shown in Table~\ref{tab:groupnumber}, 
8 groups perform comparably well for all selection ratios, and we pick 0.5 as the default. 
Smaller ratios miss the entities encoded in remaining groups while larger ratios introduce entity-irrelevant information ({\em e.g.}, background).
However, while increasing the group number to 16, it brings over-segmentation for large objects, deteriorating the performance. 
(2) the objective for cross-image consistency. 
We study another variant of cross-image consistency by directly aligning two sets of group tokens encoded with shared entity. 
We adopt the contrastive loss (NCE in Table~\ref{tab:objective}) to pull two sets of group tokens closer, while other group tokens in each mini-batch are pushed farther. Table~\ref{tab:objective} reveals the superiority of our proposed mask consistency over group consistency, which we believe is because mask consistency involves the image content to realize visual invariance.




\vspace{2pt}
\noindent\textbf{Performance on Unseen Entities.} 
One might question whether the performance gain is mainly attributed to the selected (seen) entities in the entity set. 
We measure the mean IoU for 65 objects within entity set ({\em e.g.}, person, bus) and 16 objects out of entity set ({\em e.g.}, frisbee, stop sign) on COCO. 
As shown in Table~\ref{tab:oracle}, the mIoU of unseen entities is comparable to that of seen entities, indicating our model retains strong open-vocabulary segmentation ability without being affected by the choice of entities in pre-training. The text encoder pre-trained on large text corpus remains its ability to encode the semantic concept of objects out of entity set.

\begin{table}
\centering
\resizebox{\columnwidth}{!}{
  \begin{tabular}{lcccccc}
     \toprule
    \multirow{2}{*}{COCO Object} & \multicolumn{2}{c}{Within entity set} & \multicolumn{2}{c}{Out of entity set} & \multicolumn{2}{c}{Total}  \\
    & \#Nums. & mIoU & \#Nums. & mIoU & \#Nums. & mIoU \\
     \cmidrule(r){1-1}\cmidrule(r){2-3}\cmidrule(r){4-5}\cmidrule{6-7}
     baseline  & - & 17.2 & - & 15.4 & - & 16.8 \\
     OVSegmentor & 65 & 24.8 & 16 & 25.9 & 81 & 25.1 \\
     \bottomrule
  \end{tabular}
}
\vspace{-5pt}
\caption{Performance on 65 objects within entity set and 16 objects out of entity set on COCO Object. We report baseline (with $\mathcal{L}_{\text{contrast}}$ only) results on corresponding entities for comparison.}
\vspace{-0.4cm}
\label{tab:oracle}
\end{table}

\section{Conclusion}
In this paper, we present OVSegmentor, a transformer-based model for open-vocabulary semantic segmentation. The model exploits web-collected image-caption pairs for pre-training without any mask annotations, and transfers to target benchmark segmentation datasets (including PASCAL VOC, PASCAL Context and COCO Object) in a zero-shot manner. The model clusters the image pixels into learnable group tokens, which are then aligned with the corresponding caption embeddings. We further devise two proxy tasks, namely masked entity completion and cross-image mask consistency, to learn entity-specific, fine-grained and visually invariant group semantics. OVSegmentor outperforms the state-of-the-art method on PASCAL VOC by using only 3\% (4M vs 134M) for pre-training, indicating the effectiveness and training efficiency of our model.

{\small
\bibliographystyle{ieee_fullname}
\bibliography{egbib}
}

\clearpage
\onecolumn
\appendix
\hypersetup{
colorlinks=true,
linkcolor=black
}
\renewcommand*\contentsname{Supplementary}
Here, we start by providing additional experimental results in Sec.~\ref{sec:exps} and give more visualization results in Sec.~\ref{sec:visualization}.
\section{Additional Experiments}
\label{sec:exps}

\subsection{Details on the Entity Set}
The constructed entity set contains 100 frequently appeared entities, including: 
people, man, men, woman, women, girl, boy, lady, kid, child, children, baby, student, bride, groom, couple, prince, princess, car, bus, truck, motorcycle, train, bicycle, boat, aeroplane, airplane, motorbike, bike, cup, bottle, bowl, knife, spoon, glass, fork, chair, table, bench, clock, laptop, light, vase, plant, remote, microwave, toaster, oven, mouse, keyboard, sofa, monitor, desk, tv, TV, couch, flower, refrigerator, house, building, hotel, handbag, umbrella, book, backpack, phone, shirt, tie, suitcase, T-shirt, bag, box, sink, bed, toilet, cat, dog, horse, bird, cow, sheep, elephant, bear, zebra, giraffe, ball, racket, skateboard, skis, snowboard, surfboard, kite, pizza, cake, apple, banana, sandwich, orange, carrot, donut. \textbf{Note that}, we exclude the word ``person'' in the entity set as CC12M~\cite{cc12m} claimed that they performed person-name substitutions to protect the privacy of the individuals in the images, specifically, all named entities of type Person ({\em e.g.}, the name of the artist) detected by the natural language APIs are replaced with ``person''. 

\subsection{Additional Ablation Studies}

\noindent\textbf{Effect of the Pre-trained Backbones.}
We show the effect of applying different unimodal/multimodal pre-trained weights for visual and textual encoders in Table \ref{tab:pretrain}, with $\mathcal{L}_{\text{contrast}}$ being adopted only. Training both encoders from scratch only achieves 28.8 mIoU on PASCAL VOC. Initialization from CLIP visual and text encoders (including the visual/textual projection heads) brings significant improvement. However, it requires 400M image-text pairs for pre-training. Besides, a potential drawback of applying CLIP pre-trained weights is that the model can easily learn the visual-text alignment while ignoring the visual grouping. In comparison, initializing the model from single-modality sources, i.e. DINO and BERT, yields better performance. This design choice requires no manual annotation as both DINO and BERT use self-supervised training.

\begin{table}[!htb]
\centering
\resizebox{0.5\columnwidth}{!}{
  \begin{tabular}{lccc}
     \toprule
     \multicolumn{2}{c}{Pre-training scheme} & \multirow{2}{*}{PASCAL VOC}&  \multirow{2}{*}{PASCAL Context} \\
     \cmidrule(lr){1-2}
     Visual Enc. & Text Enc. &  &  \\
     \midrule
     \XSolidBrush & \XSolidBrush & 28.8 & 12.1 \\
     CLIP-V & CLIP-T  & 38.4 & \textbf{15.3} \\
     DINO & BERT  & \textbf{40.5} & 15.1 \\
     \bottomrule
  \end{tabular}
}
\caption{Comparison of different unimodal/multimodal pre-training schemes for visual encoder and text encoder.}
\label{tab:pretrain}
\end{table}

\vspace{3pt}
\noindent\textbf{On the Choice of Mask Threshold.}
Here, we study the influence of different mask thresholds $\delta$ as mentioned in Sec.3.2 in the manuscript. As observed in Table~\ref{tab:delta}, our model reaches a decent mIoU of 53.6 on PASCAL VOC when $\delta$ is 0.6, while smaller thresholds lead to false-positive pixels of the objects. 

\begin{table}[!htb]
\centering
\resizebox{0.5\textwidth}{!}{
  \begin{tabular}{lccccccc}
     \toprule
     $\delta$& 0.3 & 0.4 & 0.5 & 0.6 & 0.7 & 0.8 & 0.9 \\
     \midrule
     mIoU & 51.2 & 51.6 & 51.8 & \textbf{53.6} & 53.1 & 53.2 & 53.4 \\
     \bottomrule
  \end{tabular}
}
\caption{Comparison of different mask thresholds $\delta$ on PASCAL VOC.}
\label{tab:delta}
\end{table}

\begin{figure}[!htb]
\centerline{\includegraphics[width=0.8\columnwidth]{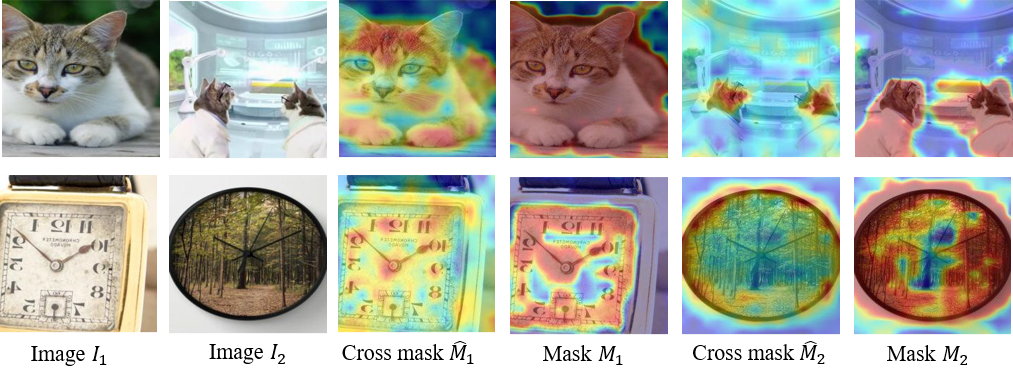}}
\vspace{-5pt}
\caption{Qualitative results of the masks and cross-image masks in our proposed cross-image mask consistency.}
\vspace{-0.2cm}
\label{fig:cross}
\end{figure}

\vspace{3pt}
\noindent\textbf{Effect of the Momentum Model.}
Our proposed OVSegmentor adopts a momentum model for encoding the cross-image, which is updated by the exponential-moving-average (EMA) of the online model. Table~\ref{tab:momentum} reveals that applying the momentum model brings about 2\% mIoU gain on PASCAL VOC and COCO Object. We attribute this to the improved quality of the pseudo targets generated by the momentum model. In Fig.~\ref{fig:cross}, we also show the object masks generated by our online model $\hat{M}_1,\hat{M}_2$ and momentum model $M_1, M_2$ for both the input image $I_1$ and the sampled cross-image $I_2$ with the shared entity.

\begin{table}[!htb]
\centering
\small
\resizebox{0.6\columnwidth}{!}{
  \begin{tabular}{lccc}
     \toprule
     Method & PASCAL VOC & PASCAL Context & COCO Object \\
     \midrule
     w/o momentum  & 50.9 & 20.2 & 23.7 \\
     w/ momentum  & \textbf{53.8} & \textbf{20.4} & \textbf{25.1}  \\
     \bottomrule
  \end{tabular}
}
\caption{Effect of the momentum model in OVSegmentor.}
\label{tab:momentum}
\end{table}

\vspace{3pt}
\noindent\textbf{Mask Probing.}
Following DINO \cite{dino} and GroupViT \cite{gvt}, we evaluate the quality of the generated masks regardless of the class predictions, termed as mask probing. Mask probing directly reflects the effect of the pixel-to-group assignment in our proposed model. For ViT-based methods that adopt finetuning transfer, {\em i.e.}, DeiT~\cite{deit}, MoCo~\cite{mocov3} and DINO~\cite{dino}, the self-attention maps in the last ViT block are probed. 

Specifically, we denote the self-attention maps as $S\in \mathbb{R}^{nh\times (L+1)\times (L+1)}$, where $nh$ refers to the number of heads and $L+1$ is the token numbers ($L$ image tokens and 1 class token), the self-attention masks $M\in\mathbb{R}^{nh\times 1\times L}$ are derived by taking the similarities of the class token and all image tokens. $M$ is then binarized by keeping the highest values ({\em e.g.} 60\%) as the foreground and the remaining regions as the background. The Jaccard similarity is computed between the attention mask for each head and the ground-truth mask, and the one with the highest similarity is taken as the mask probing result. For grouping-based methods GroupViT~\cite{gvt} and our proposed OVSegmentor, the pixel-to-group affinity $\mathbb{A}\in\mathbb{R}^{HW\times K}$ is considered as the attention masks. We directly choose one of the $K$ groups that has the highest Jaccard similarity to the ground-truth mask. As demonstrated in Table~\ref{tab:probing}, OVSegmentor surpasses methods using finetuning transfer. Despite comparable mask probing performance to GroupViT, our proposed OVSegmentor still outperforms GroupViT in terms of open-vocabulary semantic segmentation, indicating that OVSegmentor learns better group-text alignment with 85\% less data (4M vs 27M) used during pre-training. 

\begin{table}[!htb]
\centering
\resizebox{0.8\columnwidth}{!}{
  \begin{tabular}{lccccc}
     \toprule
     Method & Pretrain dataset & Supervision & Zero-shot transfer & Mask probing & Open-vocabulary segmentation \\
     \midrule
     DeiT \cite{deit} & ImageNet-1K & class & \XSolidBrush & 24.6 & 53.0 \\
     MoCo \cite{mocov3} & ImageNet-1K & self & \XSolidBrush & 28.2 & 34.3 \\
     DINO \cite{dino} & ImageNet-1K & self & \XSolidBrush& 45.9 & 39.1 \\
     DINO \cite{dino} & CC12M+YFCC15M & self &\XSolidBrush & 41.8  & 37.6 \\     
     GroupViT \cite{gvt} & CC12M+YFCC15M & text & \CheckmarkBold & 51.8 & 51.2 \\
     GroupViT* \cite{gvt} & CC4M & text & \CheckmarkBold & 45.2 & 25.8 \\
     OVSegmentor & CC4M & self+text &\CheckmarkBold & \textbf{50.9} & \textbf{53.8} \\
     \bottomrule
  \end{tabular}
}
\caption{Comparison of mask probing results on PASCAL VOC. The results of DeiT, MoCo, DINO and GroupViT are reported in \cite{gvt}.}
\label{tab:probing}
\end{table}

\vspace{3pt}
\noindent\textbf{Per-class Segmentation Performance.} 
We compare the mIoU over total 20 object categories in PASCAL VOC, as shown in Table~\ref{tab:perclass}. Our proposed OVSegmentor surpasses VIL-Seg~\cite{vilseg} on all the categories, while significantly outperforming GroupViT~\cite{gvt} on categories such as aeroplane, car, motorbike. OVSegmentor achieves inferior results on the ``person'' class, owing to its large variation of visual appearance in web-collected images, posing additional challenges for our proposed cross-image mask consistency to learn visual invariance.

\begin{table}[!htb]
\centering
\resizebox{\columnwidth}{!}{
  \begin{tabular}{lccccccccccccccccccccc}
     \toprule
     Method & Pretrain & aeroplane & bicycle & bird & boat & bottle & bus & car & cat & chair & cow & table & dog & horse & motorbike & person & plant & sheep & sofa & train & monitor \\
     \midrule
     VIL-Seg \cite{vilseg} & 12M & 40.2 & 21.6 & 41.0 & 17.3 & 35.3 & 52.8 & 10.1 & 59.3 & 15.4 & 42.4 & 21.4 & 49.5 & 56.1 & 49.4 & 11.3 & 21.6 & 41.5 & 18.6 & 54.1 & 13.4 \\
     GroupViT \cite{gvt} & 27M & 38.3 & 31.4 & 50.6 & 31.9 & \textbf{63.2} & \textbf{78.8} & 65.1 & \textbf{79.2} & \textbf{18.1} & \textbf{74.0} & \textbf{30.9} & \textbf{76.2} & 59.3 & 55.0 & \textbf{44.1} & 40.9 & 66.6 & \textbf{31.5} & 49.6 & 29.8 \\
     OVSegmentor & 4M & \textbf{ 70.8} & \textbf{32.8} & \textbf{57.5} & \textbf{40.2} & 57.3 & 76.7 & \textbf{71.7} & 77.4 & 16.5 & 72.7 & 28.2 & 61.4 & \textbf{60.0} & \textbf{70.4} & 17.8 &\textbf{ 43.3} & \textbf{69.7} & 31.2 & \textbf{58.7} & \textbf{33.2 }\\
     
     \bottomrule
  \end{tabular}
}
\caption{Comparison of per-class mIoU results on PASCAL VOC.}
\label{tab:perclass}
\end{table}








\section{More Visualization Results}
\label{sec:visualization}
\begin{figure*}[t]
\centerline{\includegraphics[width=.8\textwidth]{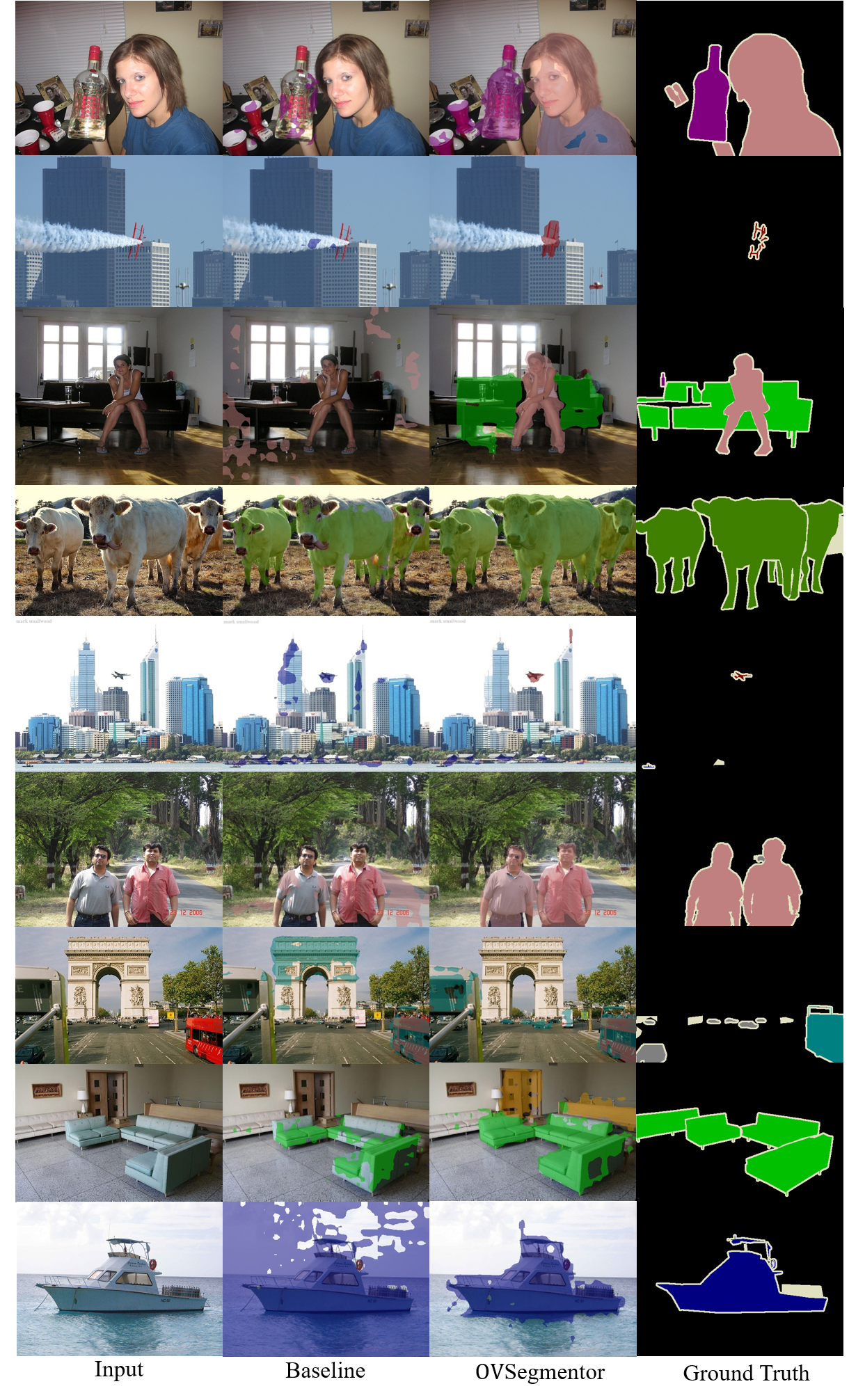}}
\vspace{-8pt}
\caption{Qualitative results on PASCAL VOC.}
\label{fig:voc}
\end{figure*}

\begin{figure*}[t]
\centerline{\includegraphics[width=.8\textwidth]{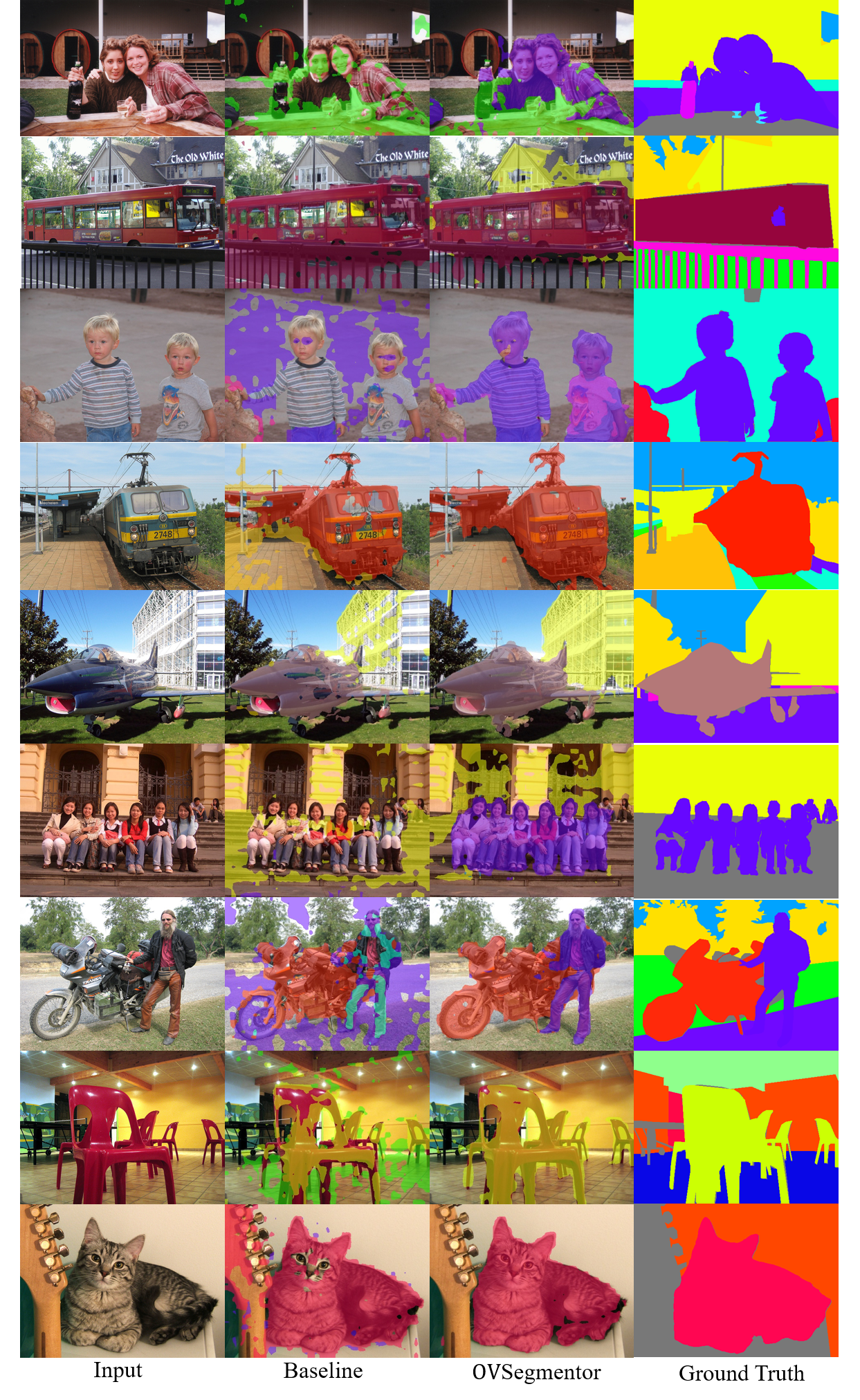}}
\vspace{-10pt}
\caption{Qualitative results on PASCAL Context.}
\label{fig:context}
\end{figure*}

\begin{figure*}[t]
\centerline{\includegraphics[width=.8\textwidth]{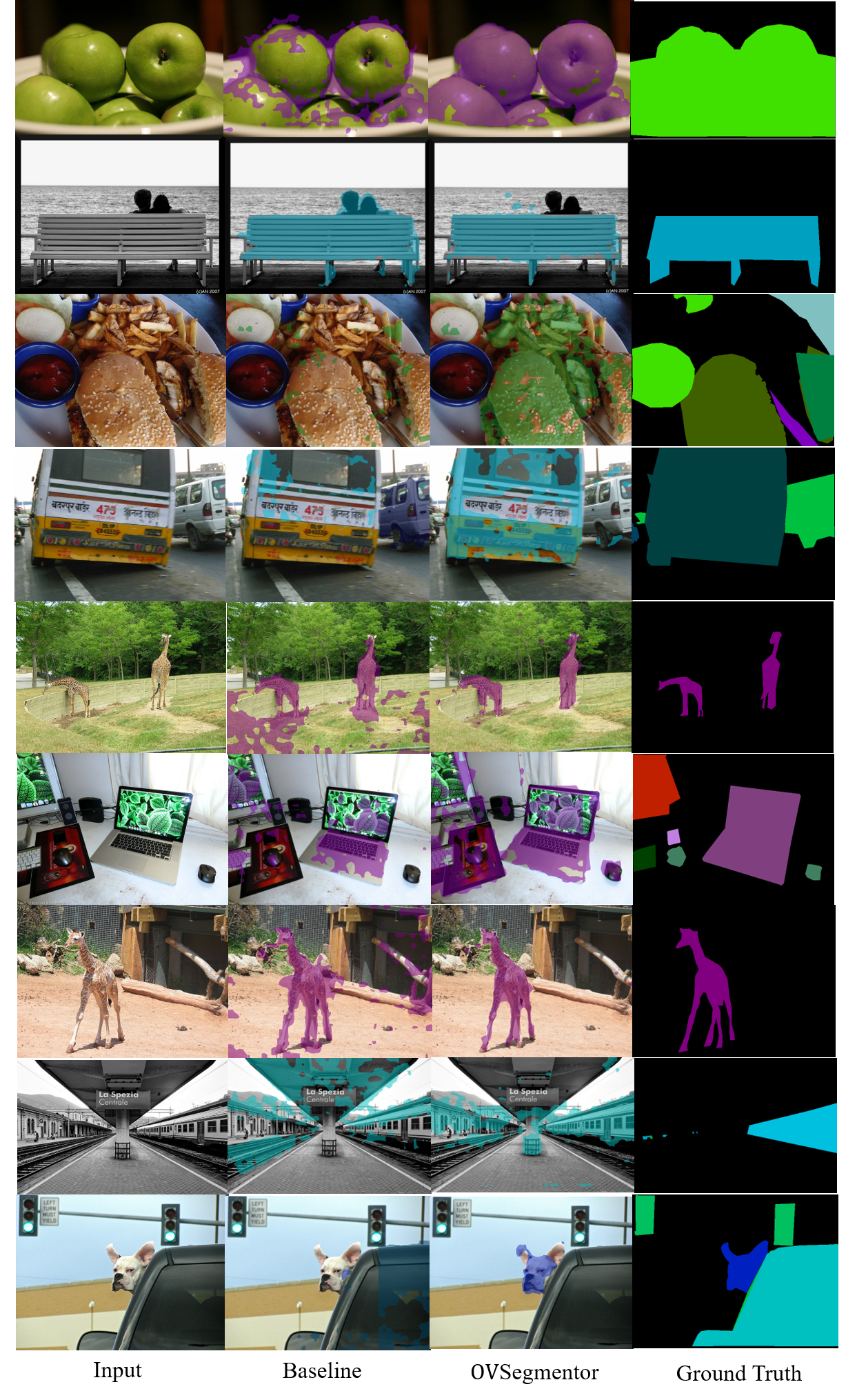}}
\vspace{-10pt}
\caption{Qualitative results on COCO Object.}
\label{fig:coco}
\end{figure*}

Additional qualitative results on PASCAL VOC, PASCAL Context, and COCO Object can be found in Fig.~\ref{fig:voc}, Fig.~\ref{fig:context}, and Fig.~\ref{fig:coco}, respectively. 
Generally, our proposed OVSegmentor successfully groups semantically related pixels together and aligns the group to the correct category. 
On PASCAL VOC, OVSegmentor successfully segments objects with various scales ({\em e.g.} small aeroplanes and distant cars in the 2$^\text{nd}$,  5$^\text{th}$ and  7$^\text{th}$ rows) and multiple objects of the same class (4$^\text{th}$, 6$^\text{th}$ and  8$^\text{th}$ rows).
In terms of PASCAL Context where objects of more categories are annotated, our model manages to segment the salient objects while failing to recognize stuff classes that usually appear as the background in web-collected data ({\em e.g.} grass, floor, wall, etc.). 
On COCO Object, we observe that our model can not separate co-occurring objects from different classes into distinct groups very well ({\em e.g.} laptop and mouse), which we conjecture is because the captions sourced from the Internet usually lack fine-grained descriptions to cover the full image content.

\end{document}